\documentclass[journal]{IEEEtran}
\usepackage{amsmath,amsfonts}
\usepackage{algorithmic}
\usepackage{algorithm}
\usepackage{array}
\usepackage{makecell}
\usepackage[caption=false,font=normalsize,labelfont=sf,textfont=sf]{subfig}
\usepackage{textcomp}
\usepackage{stfloats}
\usepackage{booktabs}
\usepackage{url}
\usepackage{times}
\usepackage{verbatim}
\usepackage{graphicx}
\usepackage{cite}
\usepackage{multirow}
\usepackage[switch]{lineno}
\usepackage{amsmath}
\usepackage{amssymb}
\usepackage{bbding}
\begin{document}
	
	\title{Mixture of Scale Experts for Alignment-free RGBT Video Object Detection and A Unified Benchmark}
	
	\author{Qishun Wang, Zhengzheng Tu, Kunpeng Wang, Le Gu, Chuanwang Guo}
	
	\markboth{Journal of \LaTeX\ Class Files,~Vol.~14, No.~8, August~2021}%
	{Shell \MakeLowercase{\textit{et al.}}: A Sample Article Using IEEEtran.cls for IEEE Journals}

	\maketitle
	
	\begin{abstract}
	Existing RGB-Thermal Video Object Detection (RGBT VOD) methods predominantly rely on the manual alignment of image pairs, that is both labor-intensive and time-consuming. This dependency significantly restricts the scalability and practical applicability of these methods in real-world scenarios. To address this critical limitation, we propose a novel framework termed the Mixture of Scale Experts Network (MSENet). MSENet integrates multiple experts trained at different perceptual scales, enabling the capture of scale discrepancies between RGB and thermal image pairs without the need for explicit alignment. Specifically, to address the issue of unaligned scales, MSENet introduces a set of experts designed to perceive the correlation between RGBT image pairs across various scales. These experts are capable of identifying and quantifying the scale differences inherent in the image pairs. Subsequently, a dynamic routing mechanism is incorporated to assign adaptive weights to each expert, allowing the network to dynamically select the most appropriate experts based on the specific characteristics of the input data. Furthermore, to address the issue of weakly unaligned positions, we integrate deformable convolution into the network. Deformable convolution is employed to learn position displacements between the RGB and thermal modalities, thereby mitigating the impact of spatial misalignment. To provide a comprehensive evaluation platform for alignment-free RGBT VOD, we introduce a new benchmark dataset. This dataset includes eleven common object categories, with a total of 60,988 images and 271,835 object instances. The dataset encompasses a wide range of scenes from both daily life and natural environments, ensuring high content diversity and complexity. Extensive experiments conducted on this benchmark dataset demonstrate the superior performance of our proposed detector compared to other state-of-the-art methods. The results highlight the effectiveness of MSENet in addressing the challenges associated with alignment-free RGBT VOD. To further promote research and development in this domain, we will publicly release our code and the benchmark dataset. 
	\end{abstract}

\section{Introduction}
\label{sec:intro}
\begin{figure}[t]  
	\centering
	\includegraphics[width=0.48\textwidth]{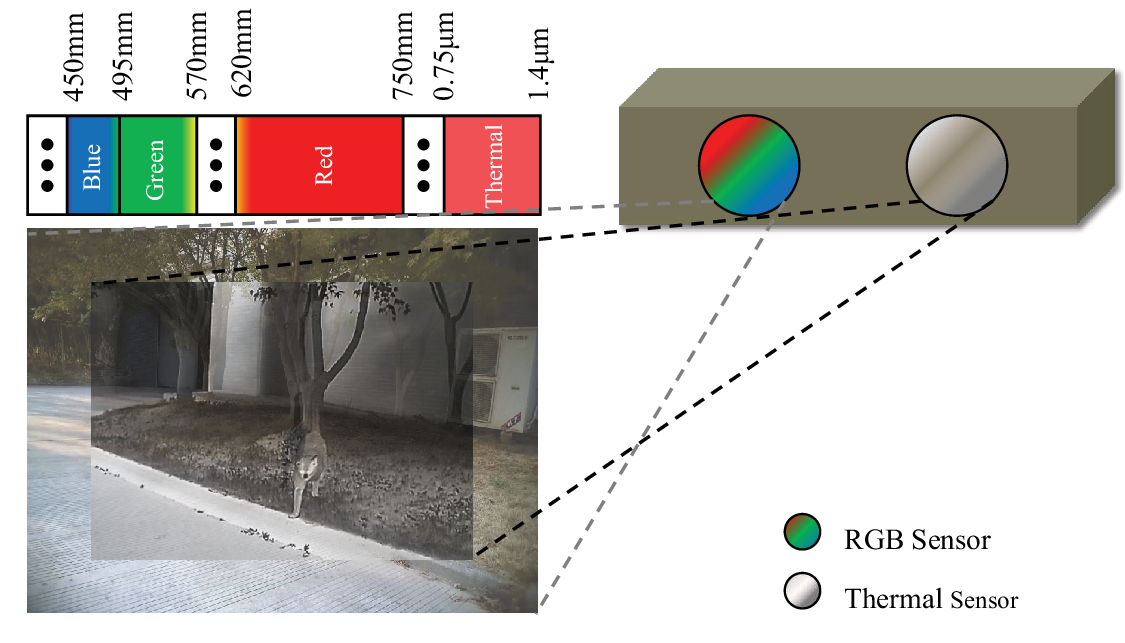}
	\caption{Multispectral sensors capture images across different wavelengths, leading to significant differences in the scale and field of view of objects. These discrepancies stem from the distinct optical properties and sensor characteristics associated with each wavelength band, resulting in variations in spatial resolution and coverage area.}
	\label{rgbtimaging}	
\end{figure}

\IEEEPARstart{T}{he} emergence of RGBT VOD \cite{tu2023erasure} represents a significant advancement over traditional RGB-based VOD systems by incorporating thermal images, which enhances detection robustness, particularly in challenging lighting conditions. Current RGBT VOD methodologies typically depend on aligned RGBT image pairs for training. However, the spatial distribution and scale of objects captured by RGB and thermal sensors differ due to variations in wavelength and focal length \cite{liu2022target}. As illustrated in Fig. \ref{rgbtimaging}, the objects in the two modalities are completely misaligned, exhibiting substantial scale variations. Even after manual adjustments to correct the object's scale, weak alignment in the RGBT image pairs persists. As a result, the manual alignment of RGBT data proves to be both challenging and time-consuming, which limits the practical application of RGBT VOD in real-world scenarios.

Currently, most RGBT vision tasks still use aligned image pairs to train multimodal networks. To address this limitation, some researchers have begun to explore how to operate on multimodal image pairs without strict alignment. Zhang \emph{et al.} \cite{zhang2021weakly} introduce the Aligned Region CNN (AR-CNN), which employs a region feature alignment module to learn positional offsets and dynamically align multimodal features corresponding to these positions within an end-to-end training framework. To handle multimodal images with significant misalignment, the method proposed by \cite{wanchaitanawong2021multi} integrates the Intersection over Union (IoU) of both modalities and utilizes a detection head for simultaneous bounding box regression across these modalities. Yuan \emph{et al.} \cite{yuan2024improving} propose a Translation-Scale-Rotation Alignment (TSRA) module designed to align region proposal features from the two modalities. However, this approach is limited to two-stage detectors. Moreover, the weakly aligned image pairs considered by the aforementioned methods are manually aligned to a similar appearance and do not faithfully represent the actual scene, as depicted in Fig. \ref{rgbtimaging}.

Recently, some studies have begun to investigate on how to use unaligned RGBT images. Song \emph{et al.} \cite{song2024misaligned} introduce a cross-modal alignment detector that focuses on learning a transformation matrix to align an RGB image with a thermal image. Similarly, Liu \emph{et al.} \cite{liu2024non} propose to estimate a homography matrix to achieve the alignment from a thermal image to an RGB image. However, there is a key challenge that semantic differences across modalities influence the reliability of the estimated transformation matrix potentially.

To address these challenges, we propose a Mixture of Scale Experts (MSE) module, designed to effectively handle the scale discrepancies and misalignments inherent in RGBT image pairs. Specifically, we configure a set of expert sub-modules, each equipped with a distinct scale perception range to capture the varying spatial characteristics of objects across modalities. These expert modules leverage the perceived differences to identify and extract the corresponding regions from the high-resolution RGB image that align with the spatial distribution of the Thermal image, thereby facilitating scale alignment. To further address pixel-level deviations between object positions in the two modalities, we employ deformable convolution. This technique dynamically adjusts sampling positions and learns positional offsets, enabling the network to adaptively correct for misalignments at the pixel level. This is particularly crucial for handling weakly aligned RGBT image pairs, where objects may exhibit significant positional discrepancies. Moreover, each expert module incorporates a scoring branch that evaluates the relevance and contribution of the expert's output to the current input. These scores are subsequently utilized as adaptive weights to dynamically fuse the results from all expert modules. This weighted fusion strategy allows the MSE to effectively adapt to RGBT image pairs with varying scale differences across diverse scenarios, thereby enhancing the robustness and accuracy of the overall system.

On the other hand, the only existing evaluation benchmark for RGBT VOD is the VT-VOD50 dataset \cite{tu2023erasure}, which comprises seven categories focused on traffic scenes and consists of 18,898 images. However, VT-VOD50 does not contain the original data but instead offers RGBT image pairs that have been manually cropped, scaled, and aligned. This deviation from the original imaging conditions limits its comprehensiveness for evaluating model performance. In response to this limitation, we propose a more realistic and extensive evaluation benchmark, named UVT-VOD2024, specifically designed for alignment-free RGBT VOD. UVT-VOD2024 encompasses a wider variety of scenes, including rural areas, towns, parks, campuses, and residential neighborhoods. With a total of 11 categories, the dataset includes 60,000 images and 271,000 object instances. In contrast to VT-VOD50, UVT-VOD2024 offers greater temporal series and scene diversity, incorporating various lighting conditions, camera motion displacements, and background complexities. Notably, it retains alignment-free RGBT image pairs derived directly from RGBT sensors. This authentic data allows for a comprehensive evaluation of RGBT VOD methods and provides insights into their potential applicability in real-world scenarios.

In summary, we have designed a Mixture of Scale Experts Network (MSENet) tailored for the alignment-free RGBT VOD task and established a comprehensive evaluation benchmark dataset, UVT-VOD2024. To the best of our knowledge, this represents the pioneering effort to explore the alignment-free RGBT VOD task. Currently, no dedicated benchmark platform exists for evaluating this specific task. The primary contributions of this study are outlined below:
\begin{itemize}
	\item We design a Mixture of Scale Experts Network (MSENet) for alignment-free RGBT VOD. Specifically, MSENet utilizes multiple experts that can dynamically perceive and eliminate scale variations across different scenes and further mitigate the misalignment of objects at the instance level through deformable convolution.
	\item We contribute a benchmark dataset for alignment-free RGBT VOD, comprising 60988 images collected from a variety of real-world scenes. The dataset is used to conduct a comprehensive evaluation and analysis of existing state-of-the-art (SOTA) detectors. It will be made publicly available.
	\item We assess the performance of our MSENet on the dataset presented within this study, confirming the efficacy of our design and attaining new SOTA results. This study provides a potentially more practical research direction for RGBT VOD.
	
\end{itemize}

\begin{figure*}[t]  
	\centering
	\includegraphics[width=0.75\textwidth]{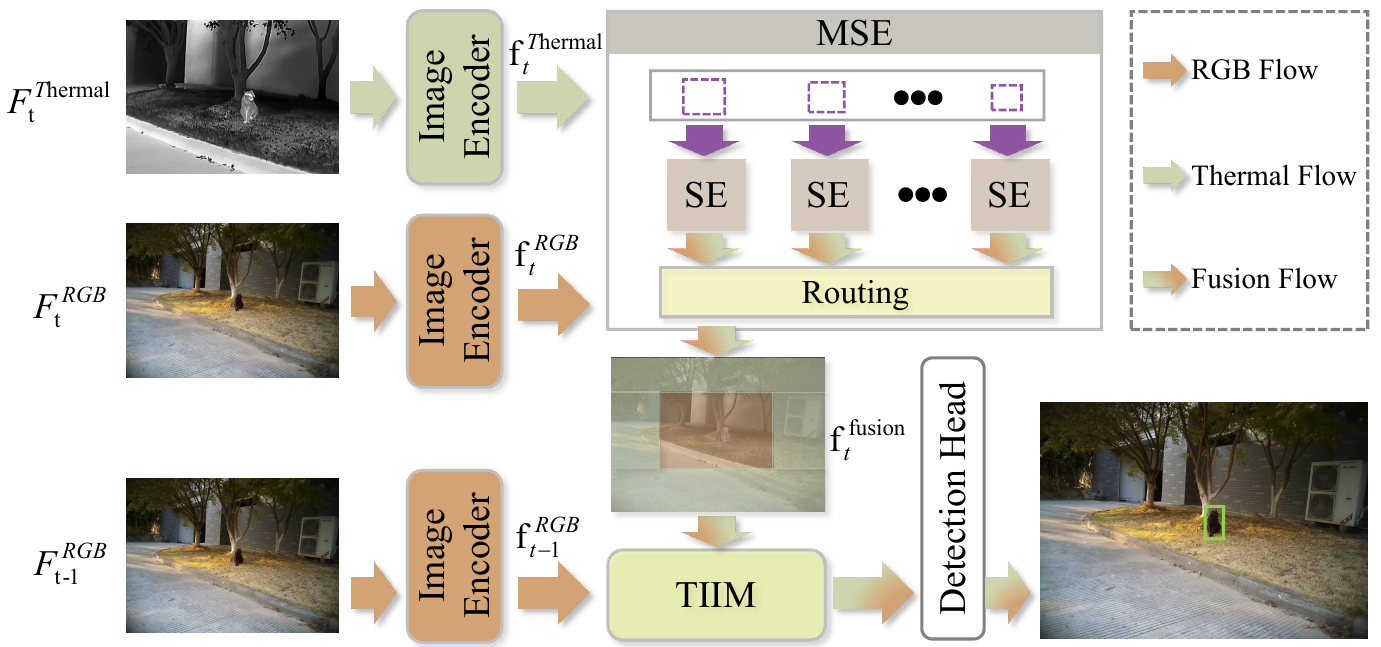}
	\caption{Architecture diagram of MSENet. Our objective is to enhance the features of the current frame $Frame_{t}^{RGB}$; therefore, we do not utilize $Frame_{t-1}^{Thermal}$ for training. This decision is based on the inherent differences in semantics, spatial context, and temporal factors between $Frame_{t}^{RGB}$ and $Frame_{t-1}^{Thermal}$. Such disparities can negatively affect the fusion effect and reduce operational efficiency.} 
	\label{net}	
\end{figure*}

\section{Related Work}
\subsection{Video Object Detection}
RGB-based VOD aims to effectively utilize temporal multi-frame information. 
Deng \emph{et al.} \cite{deng2019relation} propose employing Relation Distillation Networks (RDN) to capture long-range dependencies among objects in videos, thereby improving detection performance. 
Inspired by the human visual system's ability to observe dynamics in videos, MEGA \cite{chen2020memory} integrates global and local information comprehensively, leading to improvements in memory utility and achieving state-of-the-art performance at the time.
He \emph{et al.} \cite{he2021end} explore the potential of DETR in the VOD field through the design of a Temporal Query Encoder (TQE) and a corresponding decoder. 
Additionally, Sun \emph{et al.} \cite{sun2022efficient} propose a departure from the conventional two-stage paradigm in traditional VOD by utilizing temporal consistency to filter background areas, thus facilitating efficient single-stage VOD.
Similarly, Shi \emph{et al.} \cite{shi2023yolov} choose to model VOD as a single-stage detection problem and perform multi-frame aggregation in the later stages of the network to reduce ineffective low-quality fusion. 
Sun \emph{et al.} \cite{sun2024mamba} argue that the prior memory structure was excessively redundant.  The Hybrid Multi-Attention Transformer (HyMAT) module \cite{moorthy2025hybrid} is proposed to mitigate erroneous feature enhancement arising from positional uncertainty. This design can be seamlessly integrated into either self-attention or cross-attention mechanisms.
Therefore, they introduce a multi-level aggregation structure utilizing a memory bank, leading to a significant reduction in computing costs.

Despite advancements, RGB-based VOD continues to face imaging constraints in challenging environments. 
Tu \emph{et al.} \cite{tu2023erasure} have recently introduced RGBT VOD, addressing this issue by utilizing negative activation functions to suppress background noise and eliminating unnecessary long-term dependencies in the temporal sequence.

\subsection{RGBT Object Detection}
The incorporation of thermal images alongside RGB images for efficient object detection has become increasingly prevalent.
Guan \emph{et al.} \cite{guan2019fusion} initially propose incorporating illumination information into the neural network training process to adaptively adjust the weights of sub-branches, thereby improving the accuracy of pedestrian localization across varied environments. 
Zhou \emph{et al.} \cite{zhou2020improving} introduce a Modality Balance Network (MBNet) to address the issue of modality imbalance in pedestrian detection. 
Likewise, generative data augmentation methods are employed for domain adaptation to mitigate modality data imbalances and improve RGBT pedestrian detection performance \cite{kieu2021robust}. 
Xiang \emph{et al.} \cite{xiang2022rgb} propose initially training the feature extractors of both modalities separately, followed by feature fusion across different scales. 
Zhang \emph{et al.} \cite{zhang2023illumination} suggest combining complementary information within and across modalities to jointly capture reliable features, thereby enhancing the foreground features crucial for object detection in each modality. However, these methods all require aligned RGBT image pairs as input, necessitating significant manual effort for data preprocessing.

Zhang \emph{et al.} \cite{zhang2019weakly} initial research into pedestrian detection using weakly aligned RGBT images. They introduce a regional feature alignment module to capture positional offsets and facilitate implicit alignment. Subsequently, they extend their work to enhance the prediction method for regional offsets and refine the definition of weak alignment, addressing issues such as object-level offsets and mismatches \cite{zhang2021weakly}. Tu \emph{et al.} \cite{tu2022weakly} propose a deep correlation network that utilizes a modality alignment module based on the spatial affine transformation, the feature-wise affine transformation and the dynamic convolution to combine weakly aligned RGBT images for detecting salient objects. However, this weak alignment primarily consists of pixel-level or instance-level adjustments, which still differ from the image-level unaligned data captured by multispectral sensors in real-world scenarios.

\section{Method}
MSENet is a one-stage video detector that allows for end-to-end training, built upon the YOLOv8 framework \cite{Jocher_Ultralytics_YOLO_2023}. Specifically, YOLOV8 employs the same data loading format as YOLOV5, which facilitates seamless expansion and integration of multimodal and time-series data into our framework. In contrast, higher versions of YOLO utilize stream-based data loading, which presents challenges in terms of framework scalability and adaptability. Additionally, our empirical evaluations revealed that, in scenarios of the UVT-VOD2024 benchmark, YOLOV8 demonstrates comparable performance and efficiency to higher YOLO (You Only Look Once) counterparts. Given these findings, YOLOV8 emerges as a suitable choice, balancing ease of implementation with robust detection capabilities.
\begin{figure*}[t]  
	\centering
	\includegraphics[width=0.78\textwidth]{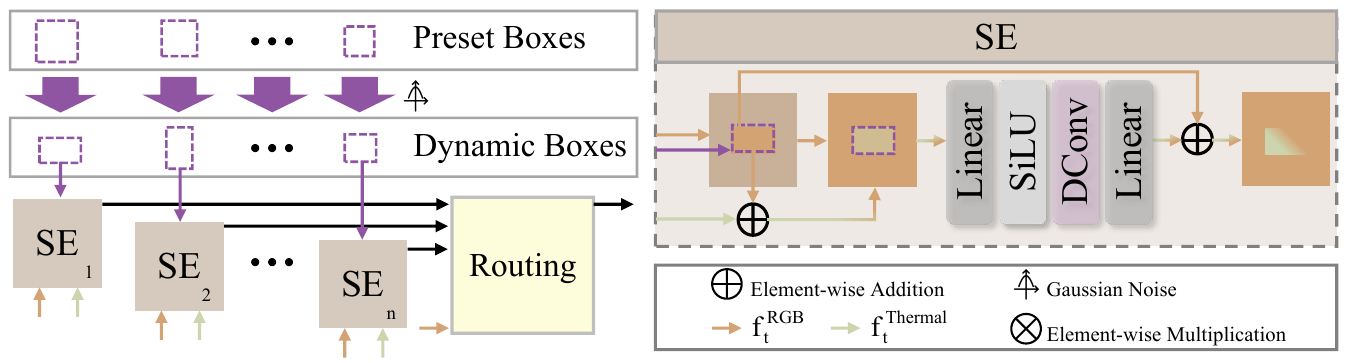}
	\caption{Flowchart of MSE module. Where ``DConv" represents the deformable convolution operation.} 
	\label{rgbt}	
\end{figure*}
\subsection{Architecture Overview}
Fig. \ref{net} illustrates the architecture of the proposed MSENet. The network is designed to accept three images as input for the training of the current frame, denoted as 
$Frame_{t}^{RGB}$. Initially, images from different modalities are processed through their respective encoders to extract features. Importantly, the encoders for different modalities do not share any learnable parameters. Once the features 
$f_{t}^{RGB}$ and $f_{t}^{Thermal}$ are processed, they are input into the MSE, where feature fusion is accomplished through collaborative perception utilizing multiple scale experts. Ultimately, after integrating temporal information, predictions are generated through the detection head.

\subsection{Image Encoder}
The image encoder employed in this work is the CSPDarkNet from the baseline method. This iteration improves upon its predecessor \cite{yolov5} by incorporating an enhanced design that includes additional skip connections and split operations. These architectural refinements not only bolster gradient flow but also significantly enhance the network's overall learning efficacy.

\subsection{Mixture of Scale Experts}
Prior fusion methodologies often assume the alignment of multi-modal images \cite{zhang2019cross,zhang2021guided,tang2022piafusion,li2023multiscale}. However, RGB and thermal images, which more accurately represent real-world scenarios, frequently exhibit considerable misalignment in scale and spatial distribution. This misalignment necessitates substantial manual effort for processing, and addressing how to circumvent this redundant step in order to enhance the applicability of RGBT VOD in practical situations has emerged as a key challenge.

Our proposed MSE utilizes a set of multi-scale mixture experts to capture the differences between RGBT image pairs at various scales. The detailed structure is illustrated in Fig. \ref{rgbt}. We have created a set of anchor boxes of various scales positioned on the $f_{t}^{RGB}$, with their dimensions determined by scale factors $\lambda$. The sizes of $f_{t}^{RGB}$ and the predefined anchor box are outlined as follows:

\begin{equation}
	f_{t}^{RGB} = (x,y,w,h)
	\label{eq1}
	\nonumber
\end{equation}
\begin{equation}
	box = (x,y,\lambda \times w,\lambda \times h),
	\label{eq1}
\end{equation}
where (x,y) denotes the center coordinates, (w,h) represents the width and height. To enhance the network's robustness against interference, we introduce a small amount of random Gaussian noise to the anchor box size.
\begin{figure}[htbp]  
	\centering
	\includegraphics[width=0.45\textwidth]{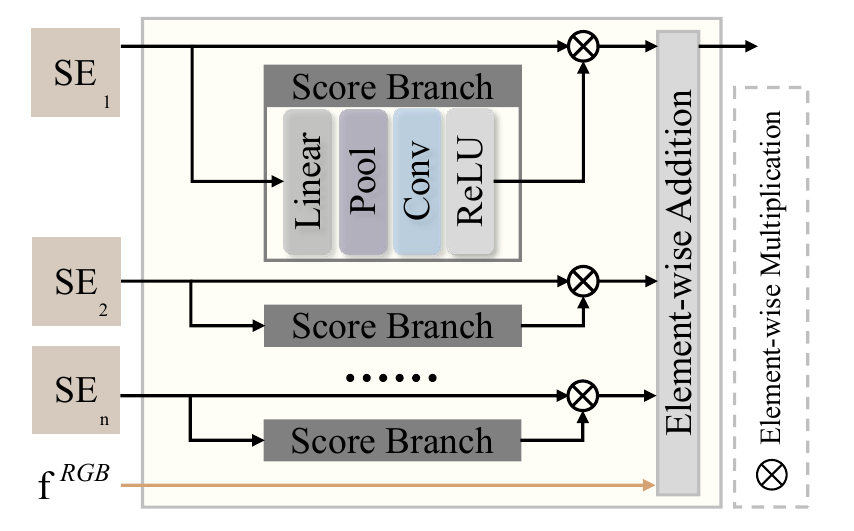}
	\caption{Flowchart of Routing in MSE.} 
	\label{route}	
\end{figure}
The quantity of anchor boxes aligns with the number of scale experts. Each expert employs an anchor box to delineate the corresponding region in $f_{t}^{RGB}$, subsequently merging it with the thermal image, thereby accomplishing preliminary alignment of the two images in spatial scale. For instance-level weak alignment of objects across different modalities, deformable convolution (referred to as DConv in Fig. \ref{rgbt}) is utilized to acquire spatial position offsets, facilitating the effective fusion of RGBT. The outcomes from all experts are consolidated through a dynamic routing mechanism.

The schematic diagram of the dynamic routing mechanism is presented in Fig. \ref{route}. It takes the outputs of the n experts and the RGB feature map as input. Each expert is associated with a score branch. To prevent bias, the learnable parameters of all score branches are shared. After scoring the fusion results from each expert, the scores are analyzed using the softmax function and employed as weights to dynamically adjust the outputs of all experts. This dynamic routing mechanism enables MSE to adapt different scale distribution discrepancies in RGBT image pairs.

\begin{figure}[htbp]  
	\centering
	\includegraphics[width=0.42\textwidth]{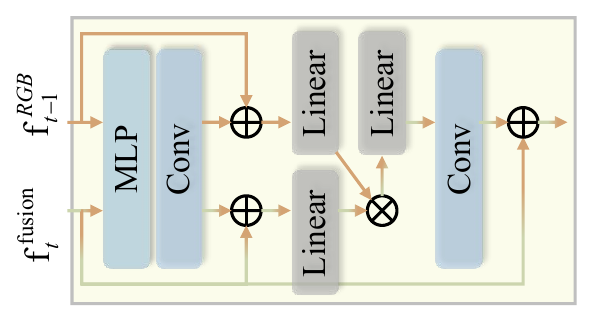}
	\caption{Flowchart of TIIM.} 
	\label{tiim}	
\end{figure}
\subsection{TIIM: Temporal Information Injection Module}
In the context of video information processing, we design a Temporal Information Injection Module (TIIM) to efficiently integrate temporal information from neighboring frames into the current frame, a process detailed in Fig. \ref{tiim}. TIIM takes inputs $f_{t}^{fusion}$ and $f_{t-1}^{RGB}$ and employs a Multi-layer Perceptron (MLP) to converge the feature embedding spaces of the two inputs, followed by a convolutional layer to capture spatial structures. Furthermore, skip connections in both branches ensure the resilience of feature learning. Subsequently, through a linear layer for mapping features, the high-dimensional feature subspace embedding is rapidly accomplished by element-wise multiplication of the feature maps from the two branches.

\subsection{Loss Function}
MSENet utilizes BCEWithLogitsLoss for computing the classification loss, which integrates both the BCELoss and Sigmoid functions to determine the loss across all category labels. The BCEWithLogitsLoss formula is as follows:
\begin{equation}
	L=-[y\cdot log(\sigma(x))+(1-y)\cdot log(1-\sigma(x))],
	\label{eq2}
\end{equation}
where $\sigma$ denotes the Sigmoid function, x represents the input value, and y signifies the target value, respectively.

For regression loss, the methods employed are Complete Intersection over Union (CIoU) \cite{zheng2021enhancing} and Distribution Focal Loss (DFL) \cite{9792391}. CIoU specifically calculates the IOU between the predicted bounding boxes and the ground truth. Ultimately, both the classification loss and regression loss are combined and fed back into the network to contribute to the parameter update process.

\begin{table*}[t]
	\centering
		\caption{Comparison between our UVT-VOD2024 and the existing dataset VT-VOD50.}
	\setlength{\tabcolsep}{1.1mm}{
		\begin{tabular}{ccccccc}
			\toprule
			Dataset & Videos & Frames & Categories  & \multicolumn{1}{l}{Instances} &Camera Movement&Scenarios \\
			\midrule
			VT-VOD50& 100   & 18898 & 7     & 202847 &\XSolidBrush &traffic\\
			\midrule
			UVT-VOD2024 (\textbf{ours})& 174   & 60988 & 11    & 271835& \Checkmark & \makecell[c]{campus, traffic, park \\ countryside, residential zone}\\
			\bottomrule
	\end{tabular}}

	\label{data-compara}
\end{table*}
\section{UVT-VOD2024: A Benchmark for Alignment-free RGBT VOD}
We have collected a dataset for alignment-free RGBT VOD to assess model performance in this area. In this section, we introduce this dataset, named UVT-VOD2024, in detail.
\begin{figure*}[t]  
	\centering
	\includegraphics[width=0.65\textwidth]{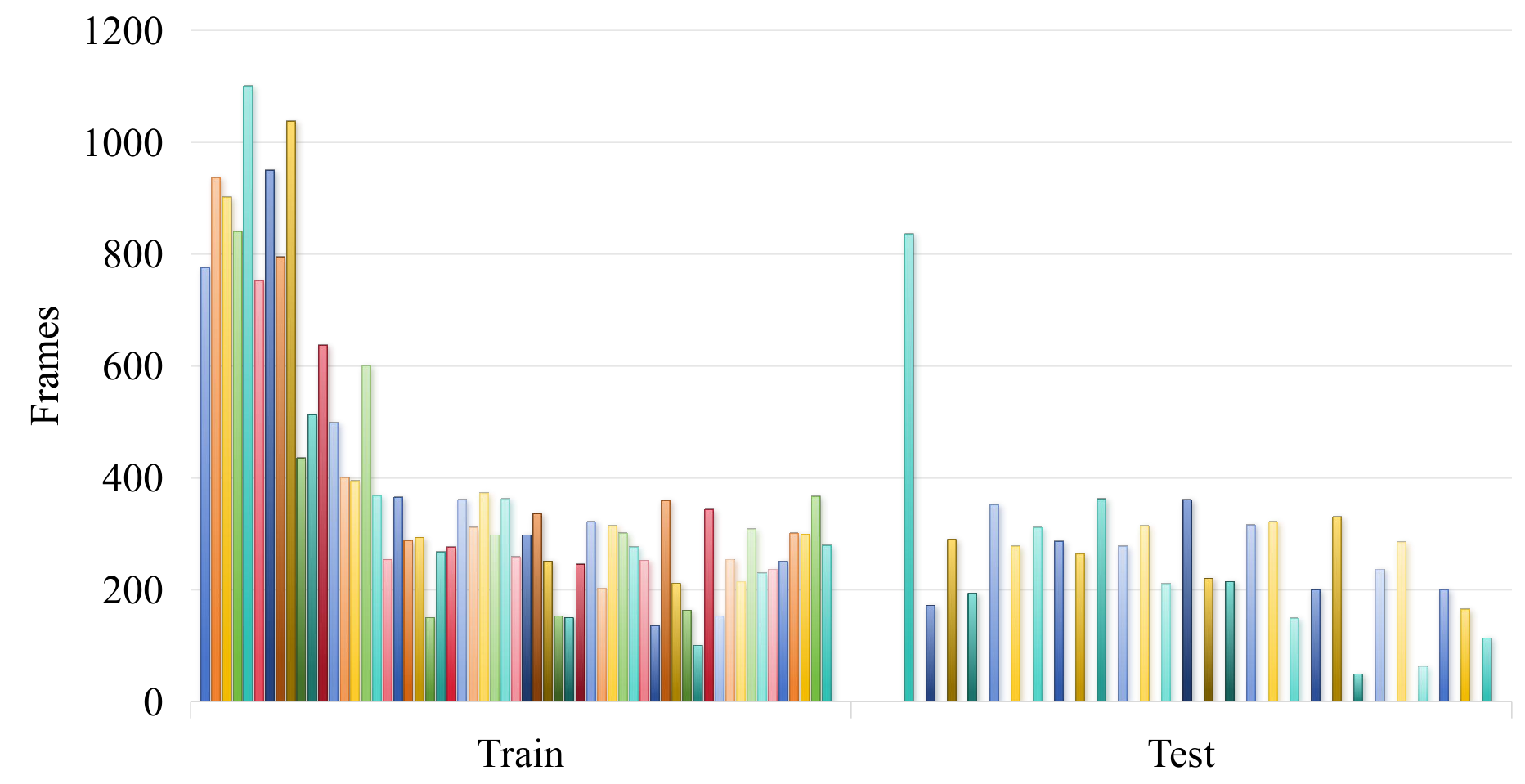}
	\caption{Training and test sets division of UVT-VOD2024. Each column represents two RGB and Thermal video pairs of exactly the same length, forming a set of multimodal data.}
	\label{trainval}	
\end{figure*}
\subsection{Data Collection and Annotation}
The equipment we utilized is the RGBT multi-spectral handheld camera by Hikvision. Data capture spans ten months, encompassing various real-life scenarios. Both modalities record video at 24 frames per second (FPS), with RGB video resolution at 1600 $\times$ 1200 and thermal video resolution at 640 $\times$ 480. 
We adopt a dynamic approach to photographing both moving and stationary objects. Data collection occurs outdoors, encompassing typical settings such as campuses, rural areas, and town roads. Subsequently, we clean data by excluding videos with blurred object definitions and severe shaking. Following this, we annotate and categorize the remaining videos.

We utilize the image annotation tool LabelImg for annotating the dataset. 
Due to resolution and scale inconsistencies between the two modalities, we adopt the annotation method employed in the RGBT VOD task, which involves annotating each frame of the RGB video with ground truth values. 
For UVT-VOD2024, we provide annotation formats in VOC \cite{Everingham10} and COCO \cite{lin2014microsoft}, which are compatible with most detection models' input requirements.
Ten students dedicate three months to completing the annotation process, with an additional month for verification to ensure consistency across annotations and mitigate potential human-induced errors.

\begin{figure*}[t]  
	\centering
	\includegraphics[width=0.85\textwidth]{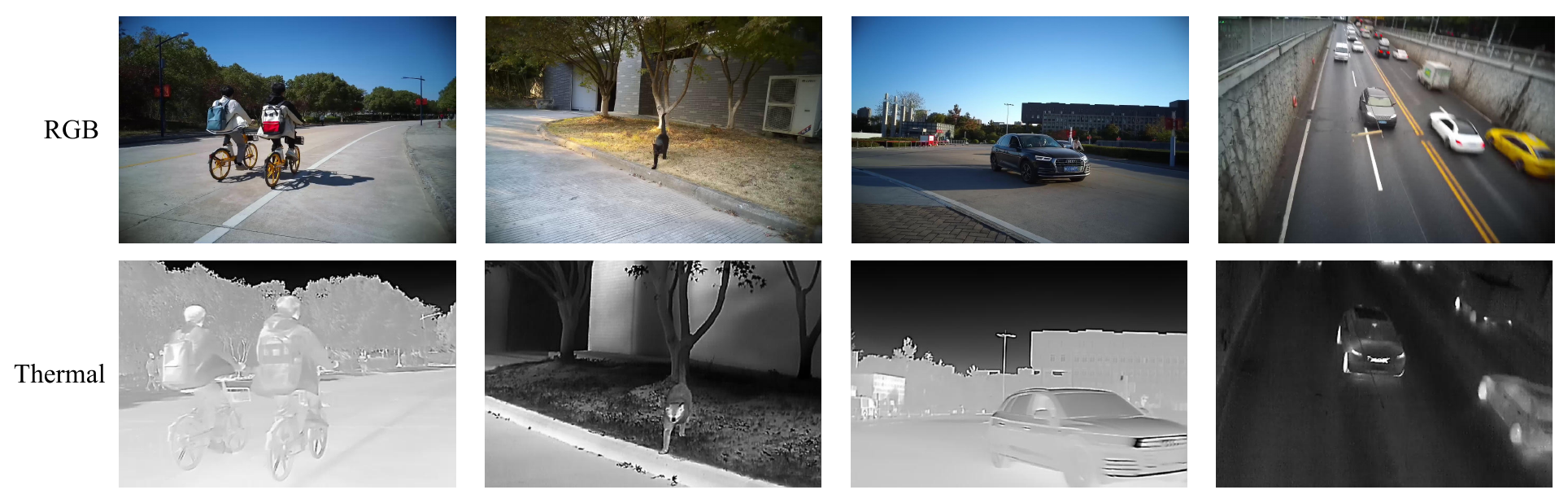}
	\caption{Examples of unaligned RGBT image pairs in UVT-VOD2024.}
	\label{data}	
\end{figure*}

\begin{figure*}[t]  
	\centering
	\includegraphics[width=0.7\textwidth]{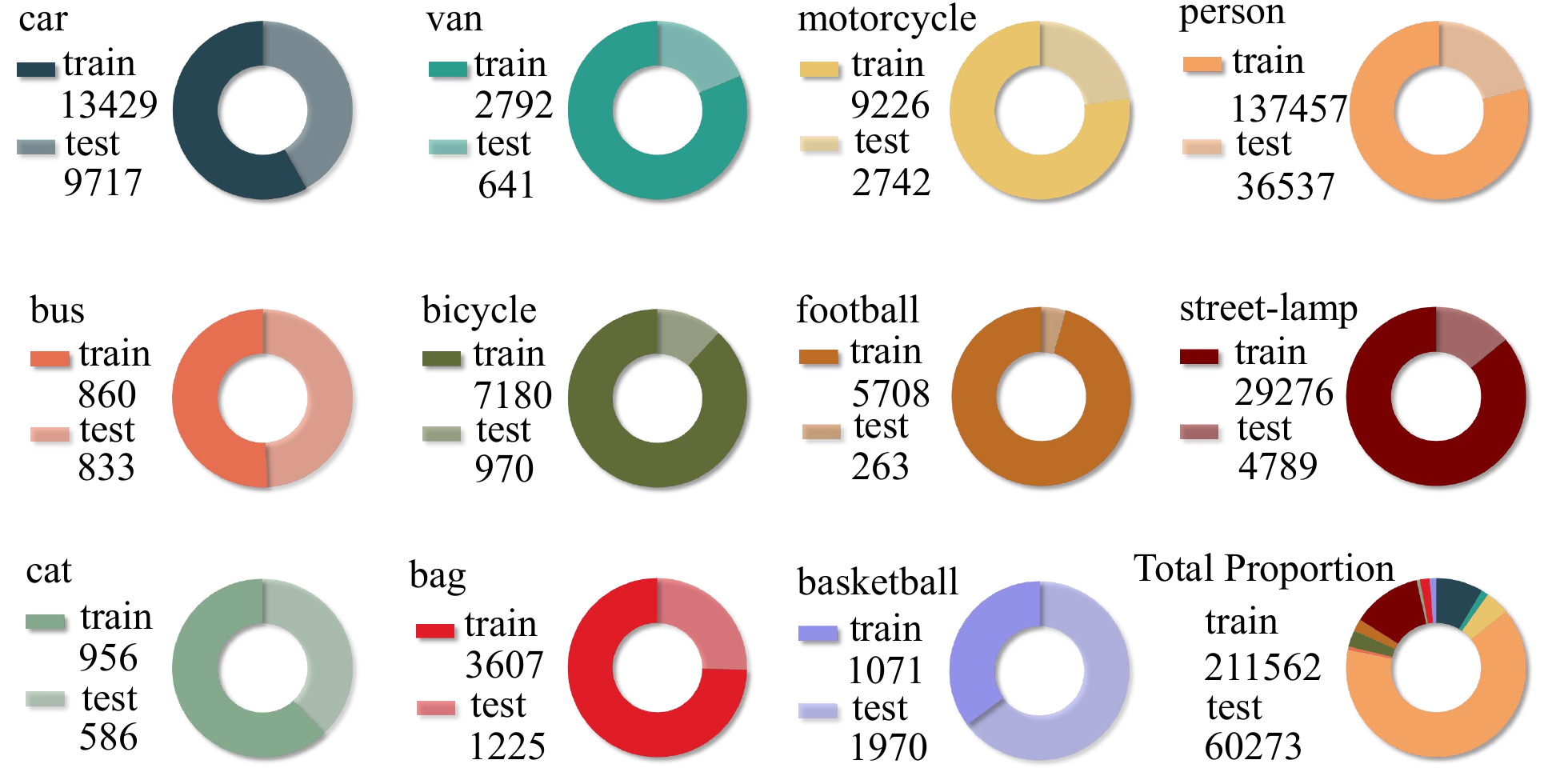}
	\caption{Details of the category and instance distribution in UVT-VOD2024.}
	\label{classes}	
\end{figure*}

\subsection{Data Description}
UVT-VOD2024 represents a significant advancement over the VT-VOD50 \cite{tu2023erasure}, in terms of both scale and scenarios. 
Detailed metrics are provided in Table \ref{data-compara}. The UVT-VOD2024 dataset comprises 174 videos of varying sizes, totaling 30,494 pairs of RGBT images. 
Of these, 118 videos are allocated for training the network, while the remaining 56 videos are reserved for evaluating its performance. 
Fig. \ref{trainval} illustrates the specific contents of both the training and test sets. 
Fig.  \ref{data} displays two pairs of unaligned raw images captured from our multispectral sensor. 
Within the UVT-VOD2024, we predefine eleven common categories of daily life scenes, with their names and distributions depicted in Fig. \ref{classes}. 
Each category contains a sufficient number of instances to enable comprehensive learning of their characteristics by the network. To more accurately reflect the complexities encountered in real-world applications, we have also introduced challenging cases with imbalanced category distributions. For instance, in the UVT-VOD2024 dataset, the instance distribution of bus is relatively sparse, the number of football instances varies significantly between the training and test sets, and basketball instances are more prevalent in the test set than in the training set. These imbalances highlight the challenges faced by object detection tasks when dealing with limited or unevenly distributed data. On one hand, these challenges effectively assess the model's performance. On the other hand, they provide a foundation for enhancing model performance, thereby increasing the likelihood of its successful application in real-world scenarios.

UVT-VOD2024 follows the classic VOC dataset format, with benchmarks and evaluation results established based on its standards, as detailed in the experimental section. 
Free access to UVT-VOD2024 is provided to facilitate public use of the dataset and ensure the reproducibility and accuracy of the research.

\begin{table*}[t]
	\centering
\caption{We evaluate MSENet and the current mainstream detection models simultaneously on UVT-VOD2024 and VT-VOD50, and we highlight the best results in \textbf{bold}. The ``-'' indicates that the measurement conditions are not met or that the result cannot be obtained.}
				\begin{tabular}{lcc|cc|cc|ccc}
					\toprule
					\multirow{2}[2]{*}{Methods} & \multirow{2}[2]{*}{Backbone} & \multirow{2}[2]{*}{Type} & \multicolumn{2}{c|}{UVT-VOD2024 (unaligned)} & \multicolumn{2}{c|}{VT-VOD50 (aligned)} & \multirow{2}[2]{*}{FPS} & \multirow{2}[2]{*}{Params(M)} & \multirow{2}[2]{*}{FLOPs(G)} \\
					&       &       & AP50(\%) & AP(\%) & AP50(\%) & AP(\%) &       &       &  \\
					
					\midrule
					YOLOV3 \cite{redmon2018yolov3}& Darknet53 & Image    &25.6   &13.5 &33.9& 17.4& 69.9 & 103.7 &283\\
					
					YOLOV5\_M \cite{yolov5} & CSPDarknet53 & Image    &  23.9 & 12.5&-&-& \textbf{294.1} & 25.1 & 64.4 \\
					
					CFT \cite{qingyun2021cross} & CFB & Image    &  6.7 &  2.4 &42.5&18.9& 222.2 & 73.7 & - \\
					
					YOLOX\_L \cite{ge2021yolox}& Darknet53 & Image     &  16.3   & - &-&-&104.8 &54.2 & 155.8 \\
					
					YOLOV6\_M \cite{li2022yolov6}  & EfficientRep & Image    & 22.7  & 12.1 &-&- & 169.5& 52 & 161.6 \\
					
					YOLOV7 \cite{10204762} & CSPDarknet53 & Image   &23 &10.4 &37.7&16.5& \textbf{294.1} & 36.5& 103.3 \\
					
					YOLOV9-C \cite{wang2024yolov9}&GELAN&Imgae& 27.3&14.5&49.1&26.9&99&25.5&103.7\\
					
					YOLOV10-M \cite{wang2024yolov10}&CSPNet&Image&17.1&8.7&46.2&25.2&210&16.5&64\\
					Efficientdet \cite{tan2020efficientdet} &EfficientNet&Image&20.2&8.8&-&-&87&20.0&100\\
					
					TOOD \cite{feng2021tood} &ResNet-50&Image&15.9&7.3&36.3&19&25.8&32&199\\
					
					Deformable DETR \cite{zhu2021deformable}&ResNet-50&Imgae&7.7&2.9&42.5&23.3&20.7&41.1&197\\
					
					RT-DETR \cite{zhao2024detrs}&ResNet-50&Imgae&17&7.9&40.2&21.6&-&42.7&130.5\\
					
					DINO \cite{zhang2022dino}&ResNet-50&Image&29.4&13.7&47.4&25.9&16.7&47.7&274\\
					
					AlignDETR \cite{cai2023align}&ResNet-50&Image&21.1&9.2&-&-&12.9&47.5&235\\
					
					DDQ DETR \cite{zhang2023dense} & ResNet-50 & Image     & 21.1  & 9.1 &48.3&26.5& 13 & 48.3& 275 \\
					
					DiffusionDet \cite{chen2023diffusiondet}&ResNet-50&Image&21.4&9.6&46.9&25.1&-&-&-\\
					\midrule
					DFF \cite{zhu2017deep}  & ResNet-50 & Video  & 9.2  & 3.9 &33.5&14.1 & 40.4 & 62.1 & \textbf{24.9} \\
					
					FGFA \cite{zhu2017flow}  & ResNet-50 & Video   & 16.7 & -  &35.1&15.8& 9 & 64.5& 41 \\
					
					RDN \cite{deng2019relation} & ResNet-50 & Video   &  16.9  &  -  &40&-& 11.3 &    -   & - \\
					
					SELSA \cite{wu2019sequence}  & ResNet-50 & Video   &  12.6 &  4.6&39.4&17.4 & 10.5&  -     & - \\
					
					MEGA \cite{chen2020memory}& ResNet-50 & Video   & 15.4  & -    &27.8&-& 16.2 &   -    & - \\
					
					Temporal ROI Align \cite{gong2021temporal} & ResNet-50 & Video   &  11.1   & 3.9  &38&17& 5.1 &    -   & - \\
					
					CVA-Net \cite{10.1007/978-3-031-16437-8_59}  & ResNet-50 & Video &  16.4    & 6.4 &39.7&19.7 & 6.9 & 41.6&548.1 \\
					
					STNet \cite{qin2023spatial}  & ResNet-50 & Video     & 15.7  & 6.5 &38.4&18.4& 5 & 41.6 & 752.3 \\
					
					EINet \cite{tu2023erasure}  & Darknet53 & Video    & 20.7& -  &46.3&24& 204.2 & \textbf{11.6}& 78.2 \\
					\midrule
					MSENet (\textbf{ours}) & CSPDarknet53 &Video   &  \textbf{33.6} & \textbf{18.7} &\textbf{50.3}&\textbf{27.6} & 131.7 &13.22 & 38.88 \\
					\bottomrule
	\end{tabular}
	
	\label{compara1}
\end{table*}

\section{Experiments}
In this section, we begin by introducing the experimental setup and foundational parameters. 
Subsequently, we compare and analyze our MSENet against existing methodologies. 
Following this, we conduct a series of ablation experiments to demonstrate the effectiveness of each component in our design.

\subsection{Datasets and Metrics}
In addition to the UVT-VOD2024 dataset introduced in this study, we also assess the model's performance using the aligned VT-VOD50 dataset \cite{tu2023erasure}, which is specifically designed for RGBT VOD. To accommodate algorithms that only process unimodal inputs, we incorporate multi-modal data by performing pixel-wise addition of the RGB and thermal images at the network input.

The evaluation criteria for the model are composed of two primary aspects. The first aspect quantifies the number of parameters and computational load, providing insight into the scale and size of the model. The second aspect evaluates model performance, where Average Precision (AP) serves as an indicator of accuracy, and Frames Per Second (FPS) reflects detection speed. Additionally, we assess parameters and the computational complexity associated with model size, presenting these findings for reference.

\subsection{Implementation Details}
The MSENet is developed using YOLOV8 \cite{Jocher_Ultralytics_YOLO_2023} architecture. 
The experiment is conducted using Python on the PyTorch framework. The network undergoes training for 100 epochs using two NVIDIA GeForce RTX 3090 GPUs, each with a batch size of 18.
The training process employs the SGD optimizer with a learning rate of 0.01 and a momentum factor of 0.9.
Default settings exclude data augmentation, utilizing solely the basic tone enhancement technique.

\subsection{Comparative Experiment}
We utilize the VT-VOD50 and UVT-VOD2024 datasets to perform a comprehensive evaluation and analysis of our proposed MSENet, alongside established mainstream detectors. The results are illustrated in Table \ref{compara1}, and a detailed analysis follows below.
\subsubsection{Results on UVT-VOD2024}
First, we examine the results on the UVT-VOD2024 dataset. In terms of the detection accuracy metric AP50, MSENet shows a 4.2\% improvement over the existing state-of-the-art method, DINO \cite{zhang2022dino}. Additionally, MSENet achieves a detection speed more than seven times faster than DINO, reaching an impressive 131 FPS. However, in terms of inference speed, MSENet takes approximately twice as long as the fastest models, YOLOv5\_M \cite{yolov5} and YOLOV7 \cite{10204762}. Despite this, it outperforms both models by an average of 10\% in AP50, reaching a value of 33.6.

Furthermore, it is observed that most image-based detectors outperform video-based detectors due to the complexity of object scenes in the UVT-VOD2024 dataset. These intricate scenes pose significant challenges for several two-stage video detectors, particularly regarding the fusion of region proposals. Additionally, the presence of unaligned features, both temporally and spatially, can disrupt the learning process within the current training framework.

Overall, the MSENet demonstrate a notable speed advantage in experiments comparing Transformer-based \cite{zhu2021deformable,zhao2024detrs,cai2023align,zhang2023dense} and Faster R-CNN-based \cite{zhu2017deep,zhu2017flow,wu2019sequence,gong2021temporal} detectors. 
However, due to the multimodal framework and multi-frame input of the MSENet, it lags behind in speed compared to the one-stage detectors represented by the YOLO series \cite{yolov5,ge2021yolox,li2022yolov6,10204762,wang2024yolov10}, but with a significant performance lead.

\subsubsection{Results on VT-VOD50}
Our proposed MSENet is effective not only for unaligned RGBT image pairs but also demonstrates robust performance on the aligned VT-VOD50 dataset. The expert configuration used in this evaluation remains unchanged and is consistent with that utilized for the UVT-VOD2024 dataset.

Among all the compared methods, YOLOv9-C \cite{wang2024yolov9} achieves the highest detection accuracy, obtaining an AP50 score of 49.1\%. 
Our MSENet outperforms YOLOv9-C by 1.2\% in accuracy and exceeds its detection speed of 99 FPS. On the VT-VOD50 dataset, Transformer-based detection methods generally perform better. In contrast, image-based detectors outperform most video-based detectors in both accuracy and speed. 

Based on the comprehensive results presented in Table \ref{compara1}, the model size and computational cost of MSENet are relatively favorable. While MSENet outperforms the comparison method in VT-VOD5, the margin is not as substantial as that observed in UVT-VOD2024. This suggests that existing models experience rapid performance degradation when faced with unaligned data, a challenge mitigated more effectively by MSENet.

\subsection{Ablation Studies}
\subsubsection{Contributions of MSE and TIIM to Proposed MSENet}
We conduct a set of ablation experiments to illustrate the development of MSENet from the baseline, as detailed in Table \ref{abla1}. 
In (a), we present the results of baseline training using RGB images alone.

\begin{table}[htbp]
	\centering
	\caption{Experimental results of different configurations in MSENet.}
	
		\begin{tabular}{cccccc}
			\toprule
			Groups&MSE&TIIM&AP50(\%)&FPS&FLOPs(G)\\
			\midrule
			(a) & & & 22.0&1111.1&8.2\\
			(b) &\Checkmark &&31.0&162.8&11.44\\
			(c) &&\Checkmark&24.0&333.3&35.65\\
			(d)&\Checkmark&\Checkmark&33.6&131.7&38.88\\
			\bottomrule
	\end{tabular}
	
	\label{abla1}
\end{table}

\subsubsection{The Impact of Different Expert Configurations in the MSE Module}
In order to investigate the impact of expert configurations on MSENet, we conducted a series of experiments, with a selection of representative results presented in Table \ref{boxes}. 
The findings reveal that the optimal performance is achieved with three groups of experts set at perceptual scales of 0.4, 0.6, and 0.8, respectively.
However, increasing the scale factors by 0.1 results in a decline in MSENet's detection performance.
Furthermore, upon increasing the number of experts to four and five, the network's performance exhibits slight fluctuations.
Notably, the inclusion of experts with smaller scale factors, such as 0.1 and 0.2, generally compromises MSENet's performance.
This decline can be attributed to the fact that scale differences of 0.1 and 0.2 do not align with the fundamental distribution laws of data in the UVT-VOD2024 dataset.

\begin{table}[t]
	\centering
	\caption{Experimental results on different expert configurations. We highlight the best results in \textbf{bold}. Here,``$\ddagger$" symbolizes the configuration employed by our proposed method.}

		\begin{tabular}{cccc}
			\toprule
			Number of Experts&Experts' Scale&AP50(\%)&AP(\%)\\
			\midrule
			1&0.3&24.2&12.8\\
			1&0.4&25.5&13.2\\
			1&0.5&25.7&13.0\\
			1&0.6&26.9&13.7\\
			1&0.7&26.3&13.2\\
			1&0.8&30.1&14.5\\
			2&0.6, 0.8&29.3&15.1\\
			2&0.4, 0.6&28.4&14.7\\
			2&0.4, 0.8&32.2&16.9\\
			$\ddagger$ 3&0.4, 0.6, 0.8&\textbf{33.6}&\textbf{18.7}\\
			3&0.5, 0.7, 0.9&31.2&15.3\\
			4&0.2, 0.4, 0.6, 0.8&29.6&16.2\\
			4&0.3, 0.5, 0.7, 0.9&32.5&17\\
			5&0.2, 0.4, 0.6, 0.8, 0.95&30.3&16.5\\
			5&0.1, 0.3, 0.5, 0.7, 0.95&30.2&15.6\\
			\bottomrule
	\end{tabular}

	\label{boxes}
\end{table}

\subsubsection{Assessment of the Structure in MSE}
To demonstrate the effectiveness of Gaussian noise and deformable convolution used in the MSE module, we conduct a set of experiments, with the results shown in Table \ref{mse}. The results demonstrate that incorporating solely the deformable convolutional layer within the MSE module yields a 2.7\% enhancement in performance. In contrast, employing only Gaussian noise results in a 1.2\% improvement. However, when these two techniques are combined, there is an additional performance boost, thereby validating the efficacy and complementary nature of both approaches.

\begin{table}[htbp]
	\centering
	\caption{Ablation study results of components in the MSE module. The results in this table are based on MSENet. We highlight the best results in \textbf{bold}.}
	\begin{tabular}{ lcc}
		\toprule
		Type&AP50(\%)&AP(\%)\\
		\midrule
		w/o Score Branch& 29.2 & 15.1\\
	w/o Gaussian Noise&31.9&16.1\\
	w/o	Deformable Convolution&30.4&15.3\\
		joint w/ MSE (\textbf{ours})&\textbf{33.6}&\textbf{18.7}\\
		\bottomrule
	\end{tabular}
	\label{mse}
\end{table}

\subsubsection{The Potential of MSE for Scene Generalization}
As previously mentioned, MSENet retains the identical expert configuration for both UVT-VOD2024 and VT-VO50 in Table \ref{compara1}. However, the scale factors ranging from 0.4 to 0.8 are notably more adept at handling unaligned RGBT images. We modify the configuration to endow an expert with a perceptual scale factor of 0.95, as presented in Table \ref{compara2}. The comparison results presented in Table \ref{compara2} are derived from the VT-VOD50 with aligned RGBT image pairs. Specifically, MSENet$^a$ denotes the configuration in Table \ref{compara1}, which comprises three experts with perception scales set at 0.4, 0.6, and 0.8. In contrast, MSENet$^b$ builds upon the MSENet$^a$ configuration by incorporating an additional expert with a perception scale of 0.95. This design choice is motivated by the fact that, objectively, a scale closer to 1 is more appropriate for scenarios involving aligned data.
\begin{table}[t]
	\centering
	\caption{Comparative experimental results of different configurations of MSENet on the VT-VOD50 dataset are presented. The configuration of MSENet$^a$ is consistent with that described in Table \ref{compara1}. In contrast, MSENet$^b$ incorporates an additional group of experts with a scale of 0.95. We highlight the best results in \textbf{bold}.}
		\begin{tabular}{lcc}
			\toprule
			Methods&AP50(\%)&AP(\%)\\
		
			\midrule
			YOLOV3 \cite{redmon2018yolov3}&33.9& 17.4\\
			
			CFT \cite{qingyun2021cross} &42.5&18.9\\
			
			YOLOV7 \cite{10204762}  &37.7&16.5\\
			
			YOLOV9-C \cite{wang2024yolov9}&49.1&26.9\\
			
			YOLOV10-M \cite{wang2024yolov10}&46.2&25.2\\
			
			TOOD \cite{feng2021tood} &36.3&19\\
			
			Deformable DETR \cite{zhu2021deformable}&42.5&23.3\\
			
			RT-DETR \cite{zhao2024detrs}&40.2&21.6\\
			
			DINO \cite{zhang2022dino}&47.4&25.9\\
			
			DDQ DETR \cite{zhang2023dense}&48.3&26.5 \\
			
			DiffusionDet \cite{chen2023diffusiondet}&46.9&25.1\\
		
			DFF \cite{zhu2017deep}   &33.5&14.1 \\
			
			FGFA \cite{zhu2017flow}  &15.8& 9  \\
			
			SELSA \cite{wu2019sequence}  &39.4&17.4 \\
			
			MEGA \cite{chen2020memory}&27.8&- \\
			
			Temporal ROI Align \cite{gong2021temporal} &38&17\\
			
			CVA-Net \cite{10.1007/978-3-031-16437-8_59} &39.7&19.7\\
			
			STNet \cite{qin2023spatial} &38.4&18.4\\
			
			EINet \cite{tu2023erasure} &46.3&24\\
			\midrule
			MSENet$^a$ (\textbf{ours}) &50.3&27.6\\
			MSENet$^b$ (\textbf{ours}) &\textbf{57.3}&\textbf{31.4}\\
			\bottomrule
	\end{tabular}
	\label{compara2}
\end{table}

The results in Table \ref{compara2} confirm our hypothesis that MSENet achieves higher detection accuracy for aligned RGBT image pairs when incorporating experts with a larger perceptual scale (0.95). Specifically, MSENet$^b$ improves AP50 by seven percentage points compared to MSENet$^a$. To further investigate the impact of different expert configurations on MSENet's performance in the aligned data scenario, we conduct additional extensive experiments, the results of which are shown in Table \ref{newexpert}.

The results in Table \ref{newexpert} demonstrate that in the scenario where RGB and Thermal images are aligned, experts with larger scales in MSENet consistently outperform those with smaller scales. Interestingly, on aligned data, relying solely on an expert with a larger perceptual scale factor yields satisfactory performance but still falls short of the performance achieved when additional experts with scales of 0.4, 0.6, and 0.8 are included. This finding appears to contradict our initial intuition. Why, then, does a smaller scale factor enhance the performance of MSENet in the aligned data scenario? Our analysis suggests that although the VT-VOD50 dataset features aligned RGBT image pairs, the entire feature map's interaction is not always necessary for all images. When the object distribution occupies a small portion of the spatial domain in some images, the network tends to focus on the local regions of the feature map. In such cases, experts with smaller perceptual scale factors play a crucial role.

In summary, the proposed MSENet not only performs well on the unaligned UVT-VOD2024 dataset introduced in this study but also adapts effectively to aligned datasets such as VT-VOD50 by adjusting the configuration of experts. This capability underscores the flexibility and robustness of MSENet.

\begin{table}[htbp]
	\centering
		\caption{The impact of different expert configurations on the performance of MSENet on the alignment dataset VT-VOD50. We highlight the best results in \textbf{bold}.}
	\begin{tabular}{ccc}
		\toprule
		Number of Experts&Experts' Scale&AP50(\%)\\
		\midrule
		4&0.4, 0.6, 0.8, 0.95&\textbf{57.3}\\
		3&0.4, 0.6, 0.8&50.3\\
		2&0.8, 0.95&55.1\\
		2&0.6, 0.95&54.7\\
		1&0.95&54.3\\
		1&0.8&52.4\\
		1&0.6&50.2\\
		\bottomrule
	\end{tabular}

	\label{newexpert}
\end{table}

\subsection{Robustness of MSENet under Different Alignment Challenges}
\begin{figure}[htbp]  
	\centering
	\includegraphics[width=0.49\textwidth]{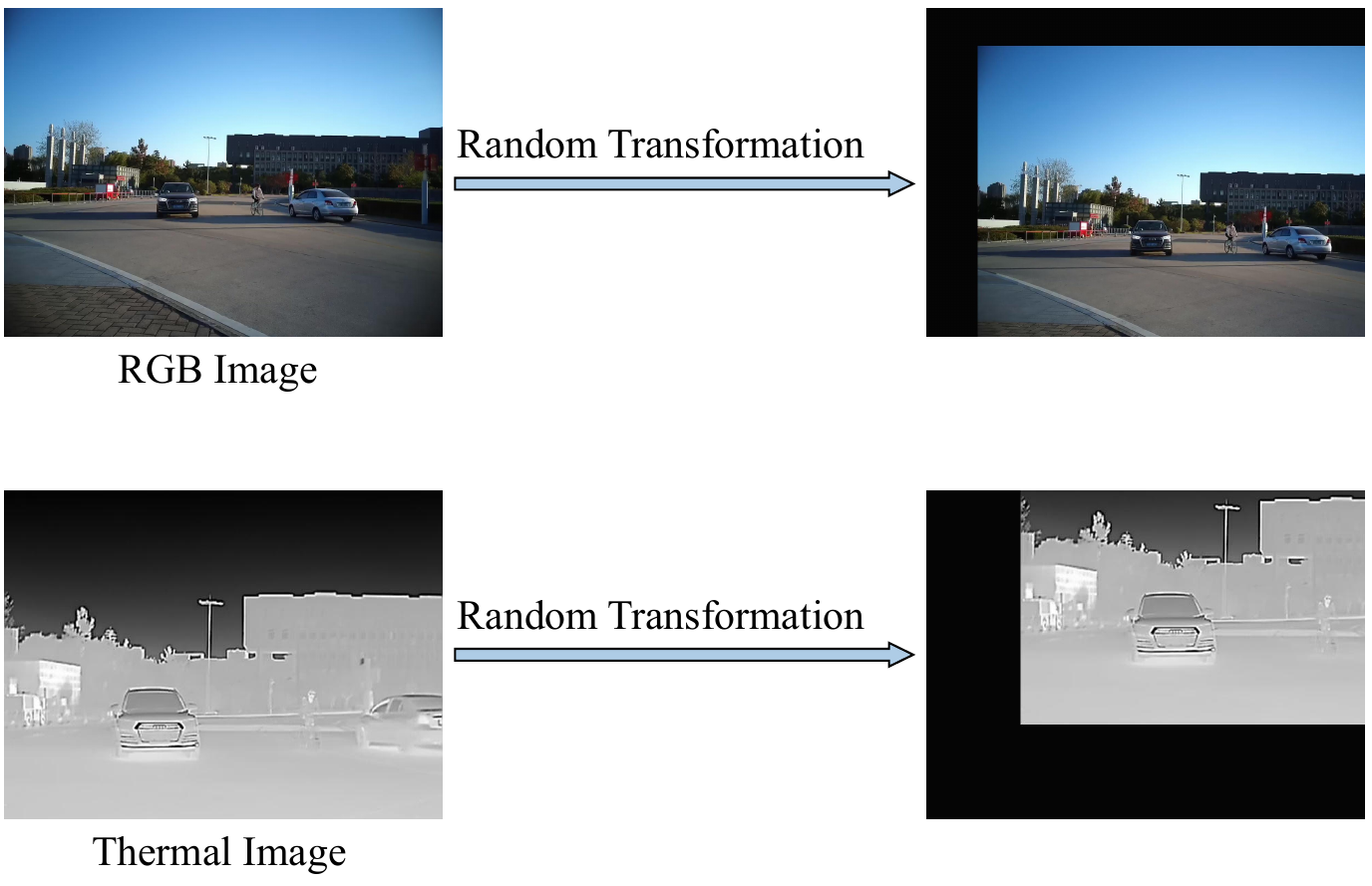}
	\caption{We employ distinct displacement parameters to conduct random spatial transformations on RGB and Thermal images separately, thereby simulating the varying degrees of spatial misalignment that may arise due to differences in camera configurations.}
	\label{viewchange}	
\end{figure}

When we introduce the UVT-VOD2024 dataset, we mention that the data are captured using a handheld Hikvision multispectral camera. For this device, the difference in viewing angle between RGB and Thermal images is nearly constant. While minor fluctuations may arise due to camera movement, the overall stability of this difference poses a unique challenge to the robustness of the network. To thoroughly assess the robustness of MSENet, we design a series of experiments. We construct a spatial transformation function designed to crop and translate the input image spatially. During training, we apply these spatial transformations independently to RGB and Thermal images. This approach enables us to simulate the varying alignments that arise when multimodal image pairs are captured by RGBT cameras with differing parameters. The above process is shown in Fig. \ref{viewchange}. We record the results of the experiment in Table \ref{viewchange_table}.

\begin{table}[htbp]
	\centering
	\caption{Effects of different ranges of spatial transformations on the performance of MSENet for RGB and Thermal images. To facilitate a comprehensive comparison of performance differences, we additionally present the performance metrics of MSENet when no spatial transformation is applied to the input images.}
	\begin{tabular}{cccc}
		\toprule
		Transformed Image&Offset Pixels &AP50(\%)&AP(\%)\\
		\midrule
		- (\textbf{ours})&0&\textbf{33.6}&\textbf{18.7}\\
		\midrule
		\multirow{3}{*}{RGB}&25 & 22 &11\\
		&50 &23.8&11.2\\
		&75&21.3&10.6\\
		\midrule
		\multirow{3}{*}{Themral}&25 &31.9  &14.7 \\
		&50 &28&12.7\\
		&75&26.9&12.5\\
		\bottomrule
	\end{tabular}
	\label{viewchange_table}
\end{table}

Table \ref{viewchange_table} demonstrates that spatial transformations of RGB images significantly impact network performance. This phenomenon arises because the training data annotations pertain exclusively to the original RGB images. Consequently, when the spatial configuration of RGB images is altered, ambiguity arises between the transformed images and their annotations, leading to a notable decline in network learning performance. In contrast, the performance degradation of Thermal images under spatial transformations is relatively minor. For instance, when the Thermal image is offset by 25 pixels, the detection accuracy decreases by only 1.7\% at an IoU threshold of 50. This resilience is attributed to the inherent characteristics of the UVT-VOD2024 dataset, which features a small thermal image perspective space that aligns well with the learning principles of MSENet. However, as the displacement pixel value in spatial transformations increases, the detection accuracy of MSENet diminishes. This is because the Thermal image provides progressively less supplementary information to the RGB image under larger displacements.

In summary, Table \ref{viewchange_table} substantiates that MSENet can achieve robust detection results even in the presence of perspective deviations, which may stem from variations in camera devices.

\section{Discussion and Future Work}
This paper represents the pioneering effort to liberate RGBT VOD tasks from the limitations of aligned data, achieving notable progress in this domain. However, MSENet remains reliant on manual expert settings, which necessitate manual adjustments in expert configurations when confronted with significant changes in data scenarios. For instance, transitioning from the unaligned UVT-VOD2024 dataset to the aligned VT-VOD50 dataset requires reconfigurations that limit the model's adaptability and scalability.

These limitations have prompted us to delve deeper into several critical issues. First and foremost is the challenge of achieving RGBT VOD without alignment while eschewing dependence on pre-defined expert settings. This involves developing more autonomous and adaptive mechanisms that can dynamically adjust to varying data characteristics without human intervention. Additionally, we aim to address more extreme scenarios, such as cases where the viewing angle of Thermal images significantly exceeds that of RGB images. These scenarios introduce unique challenges, such as increased misalignment and potential loss of complementary information, which current methods struggle to handle effectively.

In our future work, we will continue to explore these directions with the goal of enhancing the robustness, adaptability, and generalizability of RGBT VOD models.

\section{Conclusion}
This paper introduces MSENet, a novel framework designed for alignment-free RGBT VOD to more accurately adapt real-world imaging conditions. MSENet employs a set of scale-sensitive experts that collectively detect spatial discrepancies between input image pairs and dynamically route to select the most appropriate expert. This mechanism enables the identification of matching regions in RGB images corresponding to Thermal images, thereby achieving coarse alignment. Subsequently, deformable convolution is utilized to address weak alignment issues between objects in RGB and Thermal images. Additionally, we have established the UVT-VOD2024 benchmark dataset for alignment-free RGBT VOD. This dataset encompasses a wide range of life scenes and diverse weather conditions, comprising 60,988 images across 11 common categories with 271,835 object instances.

While MSENet effectively handles both aligned and unaligned data, its performance partially relies on predefined manual rules. In future work, we aim to develop more resilient and universal detectors that can operate without such dependencies, thereby enhancing the robustness and adaptability of alignment-free RGBT VOD models.

	\bibliographystyle{IEEEtran}
	\bibliography{IEEEabrv,main}
\end{document}